%% file: root.tex
\documentclass[letterpaper, 10 pt, conference]{ieeeconf}  %

\IEEEoverridecommandlockouts                              %

\usepackage{graphics} %
\usepackage{graphicx}
\usepackage{subcaption}

\usepackage{bm}

\usepackage[noadjust]{cite}

\usepackage{hyperref}
\usepackage{amsmath} %
\usepackage{amssymb}  %
\usepackage[ruled,linesnumbered,vlined]{algorithm2e}
\usepackage{booktabs}
\usepackage{multirow}
\usepackage{cleveref}
\usepackage{color, soul}

\title{\LARGE \bf
  Tightly Coupled 3D Lidar Inertial Odometry and Mapping
}

\author{Haoyang~Ye$^{1}$, Yuying~Chen$^{1}$ and~Ming~Liu$^{1}$%
\thanks{$^{1}$Haoyang Ye, Yuying Chen and Ming Liu are with RAM-LAB \url{https://ram-lab.com/}, in the Department of Electronic and Computer Engineering, Hong Kong University of Science and Technology, Kowloon, Hong Kong {\tt\small {hy.ye}@connect.ust.hk}, {\tt\small{ychenco,eelium}@ust.hk}}%
}

\begin{document}

\maketitle
\thispagestyle{empty}
\pagestyle{empty}

\begin{abstract}

Ego-motion estimation is a fundamental requirement for most mobile robotic applications.
By sensor fusion, we can compensate the deficiencies of stand-alone sensors and provide more reliable estimations.
We introduce a tightly coupled lidar-IMU fusion method in this paper.
By jointly minimizing the cost derived from lidar and IMU measurements, the lidar-IMU odometry (LIO) can perform well with acceptable drift after long-term experiment, even in challenging cases where the lidar measurements can be degraded.
Besides, to obtain more reliable estimations of the lidar poses, a rotation-constrained refinement algorithm (LIO-mapping) is proposed to further align the lidar poses with the global map.
The experiment results demonstrate that the proposed method can estimate the poses of the sensor pair at the IMU update rate with high precision, even under fast motion conditions or with insufficient features.
\end{abstract}

\setlength{\textfloatsep}{0.5pt}

\input{sections/1_introduction}

\input{sections/2_related_work}

\input{sections/3_formulas}

\input{sections/4_method}

\input{sections/5_results}

\section{Conclusion} \label{sec:conclusion}
A novel tightly coupled lidar-IMU fusion method was presented.
It comprised the state optimization for the odometry and the refinement with rotational constraints.
The results showed that our method outperformed the state-of-the-art lidar-only method and loosely coupled methods.

Despite the limitation that the proposed method requires initialization, our method indeed showed robust pose estimations results with fast update rate, even under challenging test scenarios, e.g. fast-motion cases, lidar-degraded cases and lidar sweeps with limited overlapping, empowered by sufficient IMU excitation.

\section*{Acknowledgements}
This work was supported by the National Natural Science Foundation of China (Grant No. U1713211); partially supported by the HKUST Project IGN16EG12 and Shenzhen Science, Technology and Innovation Comission (SZSTI) JCYJ20160428154842603, awarded to Prof. Ming Liu.

\bibliographystyle{IEEEtran}
\bibliography{root_bib}

\end{document}

%% file: sections/1_introduction.tex
\section{Introduction} \label{sec:introduction}

Ego-motion estimation plays a major role in many navigation tasks and is one of the key problems for autonomous robots.
It offers the knowledge of robot poses and can provide instant feedback to the pose controllers.
Besides, together with the various sensors perceiving the environment, it provides crucial information for simultaneous localization and mapping (SLAM).
Accurate estimation of the robot pose helps to reduce risks and contributes to successful planning.

The lidar sensor that can provide distance measurements for surrounding environments has been widely used in robotic systems. 
To be specific, a typical 3D lidar can sense the surroundings at a frequency around 10Hz with a horizontal field of view (FOV) of 360 degrees.
Besides, as an active sensor, it is invariant to the illumination.  
The high reliability and precision make the lidar sensor a popular option for pose estimation.

Despite its advantages, the lidar sensor is not perfect and with several shortcomings. %
From the lidar itself, it has a low vertical resolution and the sparse point cloud it obtains provides limited features, thus making feature tracking an intractable problem.
In addition, lidars mounted on moving robots suffer from motion distortion, which directly affects sensing accuracy.
In real-world scenarios, there are some lidar-degraded cases in which the lidar receives few or missing points.
For example, 3D lidar receives only few usable points in narrow corridor environments.
The received points are mainly from the side walls and only a small portion of points are observed from the ceiling and floor.
In this case, the matched lidar features can easily lead to ill-constrained pose estimation.
Another shortcoming is its low update rate.
This limits its application for tasks that require fast response such as the control of a robot pose. 

In this paper, we present a tightly coupled 3D lidar-IMU pose estimation algorithm to overcome the aforementioned problems.
Measurements from both the lidar and IMU are used for a joint-optimization.
To achieve real-time and more consistent estimation, fixed-lag smoothing and marginalization of old poses are applied, followed by a rotation-constrained refinement.
The main contributions of our work are as follows:
\begin{itemize}
\item A tightly coupled lidar-IMU odometry algorithm is proposed.
  It provides real-time accurate state estimation with high update rate.
\item  Given the prior from the lidar-IMU odometry, a rotational constrained refinement method further optimizes the final poses and the generated point-cloud maps.
  It ensures a consistent and robust estimation, even in some lidar-degraded cases.
\item The algorithm is verified with extensive indoor and outdoor tests.
  It outperforms the state-of-the-art lidar-only or loosely coupled lidar-IMU algorithms.
\item The source code is available online \footnote{\url{https://sites.google.com/view/lio-mapping} or \url{https://ram-lab.com/file/hyye/lio-mapping}}.
  This is the first open-source implementation for tightly coupled lidar and IMU fusion available to the community.
\end{itemize}

The remainder of the paper is organized as follows. Sec. \ref{sec:related_work} presents a review of related works. Sec. \ref{sec:formulas} explains the notation and some preliminaries. The proposed odometry and refinement methods are presented in Sec. \ref{sec:odom} and \ref{sec:mapping}, respectively. Implemetation and tests are shown in Sec. \ref{sec:Implemetation} and \ref{sec:results}.
Conclusion is presented in Sec. \ref{sec:conclusion}.

%% file: sections/2_related_work.tex
\section{Related Works} \label{sec:related_work}

There are several methods relating to the fusion of IMU and lidar measurements.
One important category is loosely coupled fusion.
Methods in this category consider the estimation of the lidar and the estimation of the IMU separately.
In \cite{zhang2014loam}, the lidar odometry with IMU assistance relied on the orientation calculated by the IMU and assumed a zero velocity when using the acceleration.
It decoupled the measurements of the lidar and IMU and mainly took the IMU as prior for the whole system, thus it could not utilize IMU measurements for further optimization.
In \cite{tang2015lidar}, a loosely coupled extended Kalman filter (EKF) was used to fuse an IMU and lidar in the 2D case, but it could not handle 3D or more complex environments.
Lynen et al. \cite{lynen2013robust} presented a modular way to fuse IMU measurements with other relative pose measurements, e.g., from a camera, lidar or even pressure sensor, by an EKF in the 3D case.
This loosely coupled method is of computational efficiency, but is less accurate than tightly coupled methods \cite{li2013real} since it takes the odometry part as a black box and does not update it with measurements from the IMU.

Tightly coupled methods are another important category.
For 2D planar motion estimation, Soloviev et al. \cite{soloviev2007tight} proposed a method that extracted and matched lines among the 2D lidar scans, where the tilted lidar was compensated by the predicted orientation from the IMU.
A Kalman filter was applied to correct the IMU states in the lidar measurement domain.
Hemann et al. \cite{hemann2016long} proposed a method to tightly couple the IMU propagation and accumulated lidar heightmap in the form of an error-state Kalman filter.
The corrections of the states are updated with the matching between the lidar heightmap and a priori digital elevation model (DEM). This method showed the ability in long-range GPS-denied navigation when the environments are known, but it could not work without prior map information.
In \cite{bosse2009continuous} and \cite{park2018elastic}, the raw measurements directly from the IMU and the predicted IMU measurements from the continuous trajectory were used for calculating the residuals to be optimized. The transition and estimation of the states were not involved in these methods, which made the systems unfeasible under fast motions even with an additional camera \cite{lowe2018complementary}.

Inspired by other visual-inertial works \cite{qin2018vins,leutenegger2015keyframe}, we design our method under tightly coupled lidar-IMU fusion.
We ``pre-integrate'' and use the raw IMU measurements with the lidar measurements to optimize the states within the whole system, which can work in laser degraded cases or when the motions are rapid.
To the best of our knowledge, ours is one of the few 3D lidar-IMU fusion algorithms that are suited for complex 3D environments.

%% file: sections/3_formulas.tex
\section{Notations and Preliminaries} \label{sec:formulas} %

\subsection{Notations}
We denote every line of measurement captured by the 3D lidar sensor as \textit{scan} $\mathcal{C}$, and denote \textit{sweep} $\mathcal{S}$ containing all the scans in one measurement.
For example, a 16-line 3D lidar contains 16 scans in one sweep.

In the following sections, we denote the transformation matrix as $\mathbf{T}^a_b \in SE(3)$, which transforms the point $\mathbf{x}^b \in \mathbb{R}^3$ in the frame $\mathcal{F}_\mathit{b}$ into the frame $\mathcal{F}_\mathit{a}$.
$\bar{\mathbf{T}}^a_b$ is the transform predicted by IMU.
$\mathbf{R}^a_b \in SO(3)$ and $\mathbf{p}^a_b \in \mathbb{R}^3$ are the rotation matrix and the translation vector of $\mathbf{T}^a_b$, respectively. The quaternion $\mathbf{q}^a_b$ under Hamilton notation is used, which corresponds to $\mathbf{R}^a_b$. $\otimes$ is used for the multiplication of two quaternions. We use $\hat{\mathbf{a}}_k$ and $\hat{\bm{\omega}}_k$ to denote the raw measurements of the IMU at timestamp $k$. The extracted features are denoted as $\mathbf{F}_a$ in the original capture frame $\mathcal{F}_a$, which can be transformed into the frame $\mathcal{F}_b$ as $\mathbf{F}^b_a$.

\subsection{IMU Dynamics}
\subsubsection{States}
The body frame $\mathcal{F}_\mathit{B_i}$ and $\mathcal{F}_\mathit{L_i}$ are the reference of the IMU body and the reference of the lidar center, respectively, while obtaining the lidar sweep $\mathcal{S}_i$ at the discrete timestamp $i$.
The states we will estimate are the IMU state $\mathbf{X}^W_{B_i}$ in the world frame $\mathcal{F}_\mathit{W}$ and the extrinsic parameters $\mathbf{T}^L_B$ between the lidar and the IMU sensors.
In detail, we can write the IMU states at $i$ and the extrinsic parameters as
\begin{equation} \label{eqn:states}
  \begin{aligned}
    \mathbf{X}^W_{B_i}&=
    \begin{bmatrix}
      {\mathbf{p}^W_{B_i}}^T & {\mathbf{v}^W_{B_i}}^T & {\mathbf{q}^W_{B_i}}^T & {\mathbf{b}_{a_i}}^T & {\mathbf{b}_{g_i}}^T
    \end{bmatrix}^T \\
    \mathbf{T}^L_B&=
    \begin{bmatrix}
      {\mathbf{p}^L_B}^T & {\mathbf{q}^L_B}^T
    \end{bmatrix}^T
  \end{aligned},
\end{equation}
where $\mathbf{p}^W_{B_i}$, $\mathbf{v}^W_{B_i}$ and $\mathbf{q}^W_{B_i}$ are the position, velocity and orientation of the body frame w.r.t. to the world frame, respectively, $\mathbf{b}_a$ is the IMU acceleration bias and $\mathbf{b}_g$ is the IMU gyroscope bias.

\subsubsection{Dynamic model}
With the inputs from the IMU's accelerator and gyroscope, we can update the preceding IMU state $\mathbf{X}^W_{B_i}$ to current IMU state $\mathbf{X}^W_{B_{j}}$ by discrete evolution, as shown in Eq. (\ref{eqn:imu_update}), where $\Delta t$ is the interval between two consecutive IMU measurements, and all the IMU measurements between the lidar sweeps' times $k=i$ and $k=j$ are integrated.
With slight abuse of notation, we use $k=j-1$ as the preceding IMU timestamp before $k=j$.

\begin{equation} \label{eqn:imu_update}
  \begin{aligned}
  &\mathbf{p}_{j} = \mathbf{p}_{i} + \sum_{k=i}^{j-1}\left[\mathbf{v}_{k}\Delta t + \frac{1}{2}\mathbf{g}^W \Delta t^2 + \frac{1}{2}\bm{R}_{k}(\hat{\mathbf{a}}_{k}-\mathbf{b}_{a_k})\Delta t^2\right] \\
  &\mathbf{v}_{j} = \mathbf{v}_{i} + \mathbf{g}^W\Delta t_{ij} + \sum_{k=i}^{j-1}\mathbf{R}_{k}(\hat{\mathbf{a}}_{k}-\mathbf{b}_{a_k})\Delta t \\
  &\mathbf{q}_{j} = \mathbf{q}_{i}\otimes \prod_{k=i}^{j-1}\delta \mathbf{q}_k = \mathbf{q}_{i}\otimes \prod_{k=i}^{j-1}\begin{bmatrix}\frac{1}{2}\Delta t(\hat{\bm{\omega}}_{k}-\mathbf{b}_{g_k}) \\ 1\end{bmatrix},
  \end{aligned}
\end{equation}
where $\mathbf{g}^W$ is the gravity vector in the world frame. We use the shorthands $(\cdot)_i \doteq (\cdot)^W_{B_i} $, $\Delta t_{ij} = \sum_{k=i}^{j-1}\Delta t$ and $\prod_{k=i}^{j-1}$ as the sequences of quaternion multiplications for clarity.

\subsubsection{Pre-integration}
The body motion between the timestamps $i$ and $j$ can be represented via pre-integration measurement $z^i_j = \left\{ \Delta \mathbf{p}_{ij}, \Delta \mathbf{v}_{ij}, \Delta \mathbf{q}_{ij} \right\}$, which has the covariance $\mathbf{C}^{B_i}_{B_j}$ in the error-state model (see details in the supplementary material \cite{ye2018supplementary}).

%% file: sections/4_method.tex
\section{Tightly coupled Lidar-IMU Odometry} \label{sec:odom}

To ensure efficient estimation, many works on lidar mapping, like \cite{bosse2009continuous}, \cite{zhang2014loam} and \cite{behley2018efficient}, separate the task into two parts, the odometry and the mapping.
Inspired by these works, the proposed system comprises two parallel parts. The first part, introduced in Sec. \ref{sec:odom}, is the tightly coupled lidar-IMU odometry, which optimizes all the states within a local window.
The second part, presented in Sec. \ref{sec:mapping}, is the rotation constrained refinement (leading to a globally consistent mapping process), which aligns the lidar sweeps to the global map using the information from the optimized poses and gravity constraints.

\subsection{Lidar-IMU Odometry Overview} \label{sec:odom_overview}

\begin{figure}[!ht]
\centering
\includegraphics[width=0.45\textwidth]{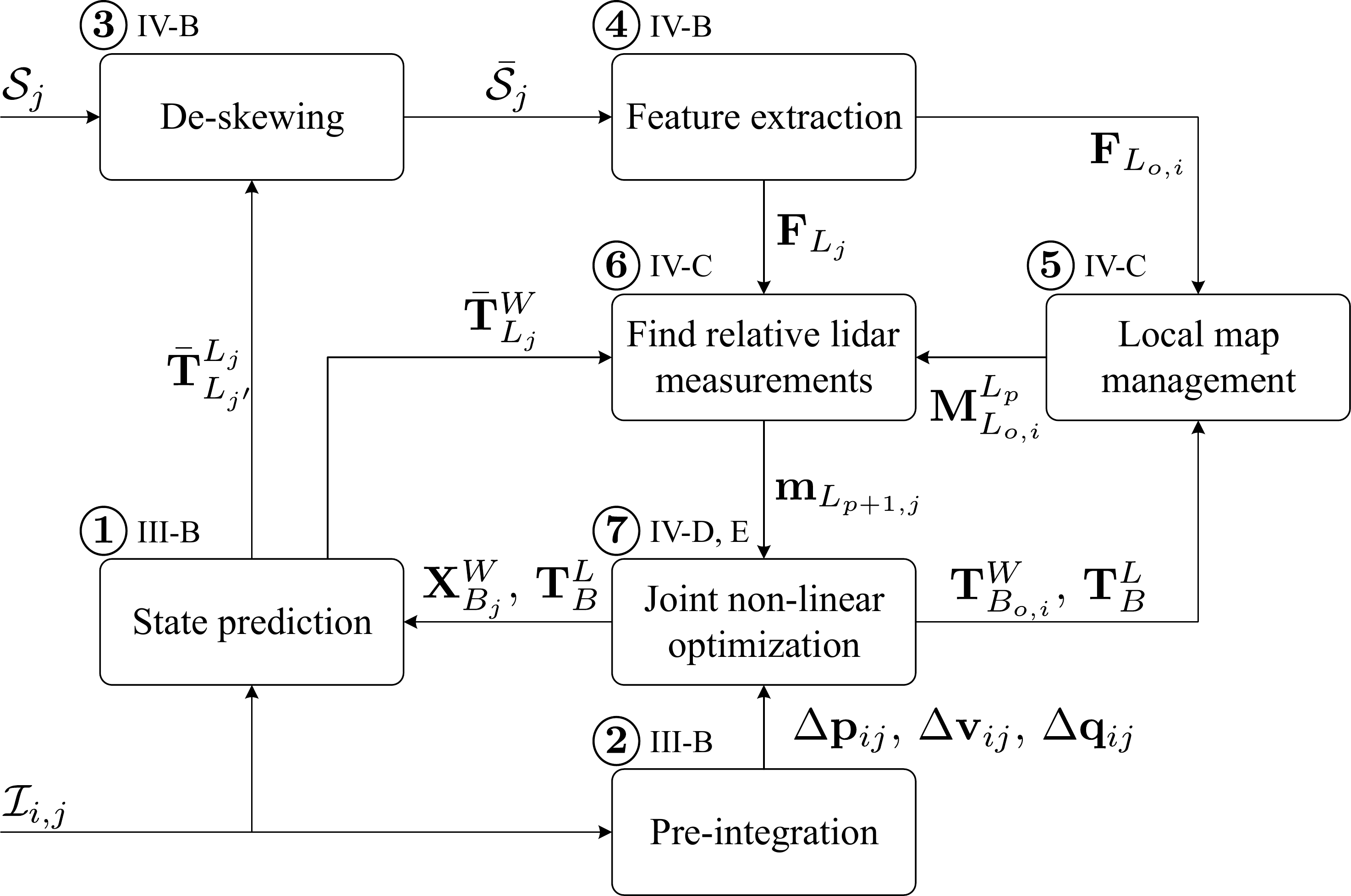}
\caption{Lidar-IMU odometry framework.
  The subsection that discussed each block is given after the circular number.
}
\label{fig:system_diagram}
\end{figure}

Fig. \ref{fig:system_diagram} provides a brief overview of our proposed lidar-IMU odometry. With the previous estimated states, we can use the current lidar raw input $\mathcal{S}_j$, and the IMU raw inputs $\mathcal{I}_{i,j}$, from the last timestamp $i$ to the current timestamp $j$, to have a new step of optimization for the states. The odometry estimation performs as follows:
  1) Before $S_{j}$ arrives, the IMU states are updated via Eq. (\ref{eqn:imu_update}) iteratively.
  2) Meanwhile, these inputs are ``pre-integrated'' as $\Delta \mathbf{p}_{ij}$, $\Delta \mathbf{v}_{ij}$ and $\Delta \mathbf{q}_{ij}$ to be used in the joint optimization.
  3) When the latest lidar sweep $\mathcal{S}_j$ is received, de-skewing is applied on the raw data to obtain the de-skewed lidar sweep $\bar{\mathcal{S}}_j$ (Sec. \ref{sec:preproc}).
  4) Next, a feature extraction step is applied to reduce the dimension of the data and extract the most important feature points $\mathbf{F}_{L_j}$ (Sec. \ref{sec:preproc}).
  5) The previous lidar feature points $\mathbf{F}_{L_{o,i}}$ within the local window are merged as a local map $\mathbf{M}^{L_p}_{L_{o,i}}$, according to the previous corresponding optimized states $\mathbf{T}^W_{B_{o,i}}$ and $\mathbf{T}^L_B$ (Sec. \ref{sec:relative_lidar}).
  6) With the predicted lidar pose for $\mathbf{F}_j$, we can find the relative lidar measurements $\mathbf{m}_{L_{p+1,j}}$ (Sec. \ref{sec:relative_lidar}).
  7) The final step is joint non-linear optimization, taking the relative lidar measurements and IMU pre-integration to obtain a MAP estimation of the states within the local window (Sec. \ref{sec:lidar_matching} and \ref{sec:optimization}). The optimized results are applied to update the prediction states in step 1), avoiding the drift from IMU propagation.

\subsection{De-skewing and Feature Extraction} \label{sec:preproc}
The 3D lidar has rotating mechanism inside to receive data for a whole circle.
When the 3D lidar is moving, $\mathcal{S}_j$, the raw data from it, suffers from motion distortion, which makes the point in a sweep different from the true positions. To handle this problem, we use the prediction of lidar motion $\bar{\mathbf{T}}^{L_j}_{L_{j'}}$ from IMU propagation and assume the linear motion model during the sweep.
Then, every point $\mathbf{x}(t) \in \mathcal{S}_j \subset \mathbb{R}^3$ is corrected by linear interpolation of $\bar{\mathbf{T}}^{L_j}_{L_{j'}}$ to obtain the deskewed sweep $\bar{\mathcal{S}}_j$ into the ending pose of the sweep,  where $t \in \left(t_{j'}, t_j\right]$ is the timestamp of the point in the sweep, and $t_{j'}$ and $t_j$ are the timstamps of the sweep start and end, respectively.

For computational efficiency, lidar feature extraction is required. Here, we are only interested in the points which are most alike on a plane or on an edge \cite{bosse2009continuous, zhang2014loam}, since these points can be extracted from every scan of a lidar sweep. Such feature points $\mathbf{F}_{L_j}$ in $\bar{\mathcal{S}}_j$ are selected by the curvature and distance changes, as are those in \cite{zhang2014loam}; i.e., the most planar or edged points are selected.

\subsection{Relative Lidar Measurements} \label{sec:relative_lidar}

\begin{figure}[tb]
\centering
\includegraphics[width=0.4\textwidth]{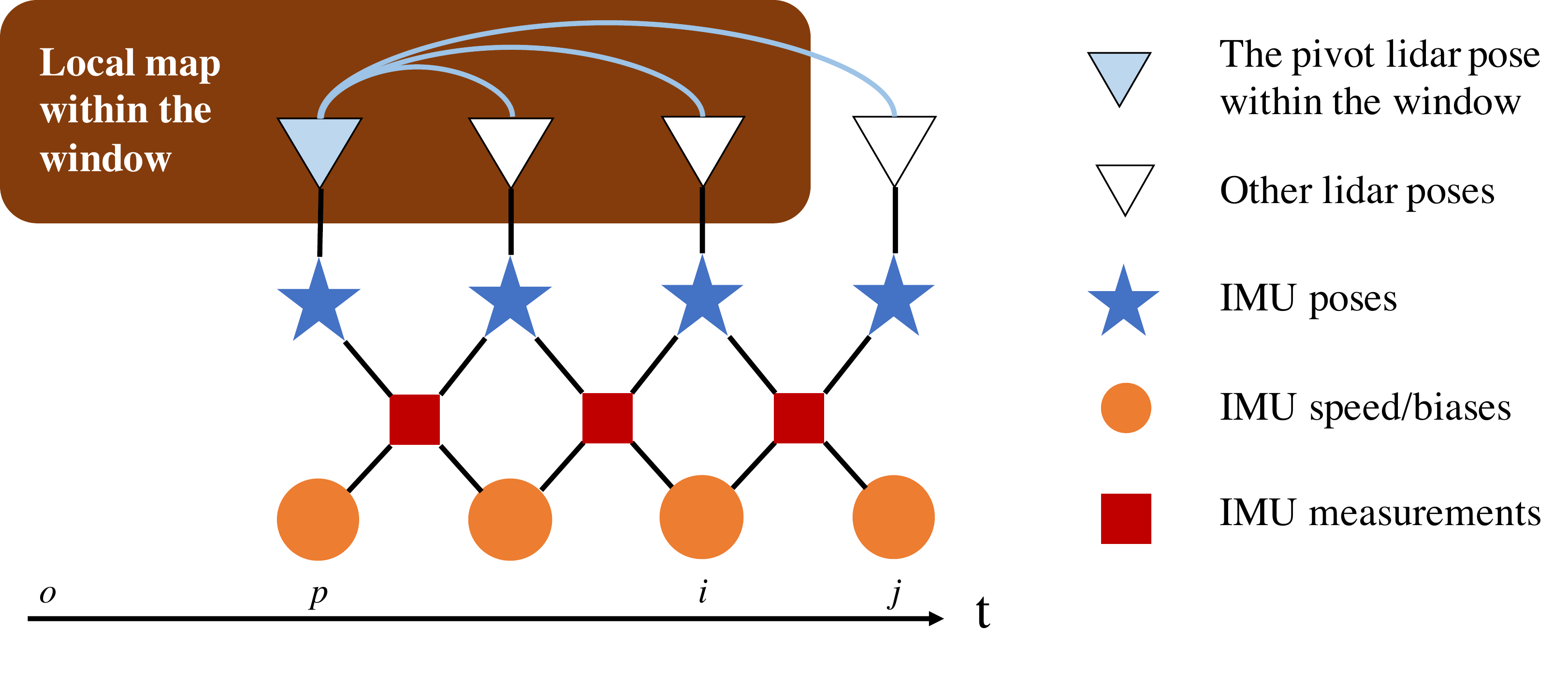}
\caption{Local window. The local map consists of the previous point-clouds before $j$ and starting from $o$. The optimization window contains the frame between $p$ and $j$.}
\label{fig:local_window}
\vspace{-0.5em}
\end{figure}

With the fusion of the IMU and another sensor, which is capable to provide the relative pose of the sensor pair, the states to be estimated, $\mathbf{X}^W_B$ and $\mathbf{T}^L_B$, will be locally observable if we fix the first reference frame \cite{jones2011visual}.
To properly incorporate pre-integration from the IMU, we propose using the relative lidar measurements, between sweeps to constrain the lidar poses, as Algorithm \ref{alg:features}.
Before finding the point correspondences, we build a local map, because the points in a single sweep are not dense enough to calculate accurate correspondences.

The local map contains the lidar feature points from $N_m$ discrete timestamps $\{o,\cdots, p, \cdots, i\}$, where $o$, $p$ and $i$ are the timestamps of the \textit{first} lidar sweep within the window, the \textit{pivot} lidar sweep and the \textit{last} processed lidar sweep, respectively, as shown in Fig. \ref{fig:local_window}.
The local map $\mathbf{M}^{L_p}_{L_{o,i}}$ is built in the frame of the $pivot$ lidar sweep from the features $\mathbf{F}^{L_p}_{L_{\gamma}}, \gamma \in \{o, \cdots, i\}$, which is transformed via the previous optimized lidar poses $\mathbf{T}^{L_p}_{L_{\gamma}}$ \footnote{For simplification, we denote the predicted transform $\bar{\mathbf{T}}^{L_p}_{L_j}$ as $\mathbf{T}^{L_p}_{L_j}$ in this section, and $\mathbf{F}^{L_p}_{L_j}$ is transformed via $\bar{\mathbf{T}}^{L_p}_{L_j}$.}.
The to-be-estimated states are the ones at the $N_s$ timestamps $\{p+1, \cdots, i, j\}$, where $p+1$ and $j$ are the timestamp of the lidar sweep next to the $pivot$ one and the current lidar sweep in the window.

\begin{algorithm}

  \SetAlgoLined
  \KwIn{$\mathbf{F}_{L_{\gamma}}$ and $ \mathbf{T}^W_{L_{\gamma}}$, $\gamma \in \{o, \cdots, i, j\}$}
  \KwOut{$\mathbf{m}_{L_{\alpha}}$, $\alpha\in \{p+1, \cdots, i, j\}$}
  \For{$\gamma \leftarrow o$ \KwTo $j$}{ \label{alg:alg1_corr_begin}
    Transform $\mathbf{F}_{L_{\gamma}}$ into $\mathcal{F}_{L_p}$ as $\mathbf{F}^{L_p}_{L_{\gamma}}$ by $\mathbf{T}^{L_p}_{L_{\gamma}}$\;
    \If{$\gamma \neq j$}{
      Merge $\mathbf{F}^{L_p}_{L_{\gamma}}$ into  $\mathbf{M}^{L_p}_{L_{o,i}}$\;
    }
  }
  \For{$\alpha \leftarrow p+1$ \KwTo $j$}{
    Find KNN($\mathbf{F}^{L_p}_{L_{\alpha}}$) in $\mathbf{M}^{L_p}_{L_{o,i}}$\;
    For each point $\mathbf{x} \in \mathbf{F}_{L_{\alpha}}$, use $\pi(\mathbf{x}^{L_p})$ to obtain the relative measurement model, which forms $\mathbf{m}_{L_{\alpha}}$\;
  }\label{alg:alg1_corr_end}

  \caption{Relative Lidar Measurements}
  \label{alg:features}
\end{algorithm}

With the built local map, the correspondences can be found between $\mathbf{M}^{L_p}_{L_{o,i}}$ and the original $\mathbf{F}_{L_{\alpha}}, \alpha \in \{p+1, \cdots, j\}$.
We define such correspondence as relative lidar measurements, since they are relative to the $pivot$ pose, and the $pivot$ pose will change with the sliding window. The original features we extracted in Sec. \ref{sec:preproc} are the most planar or edged points in $\mathcal{F}_{L_{\alpha}}$.
In practice, we found that the edged points cannot improve the results of the lidar-IMU odometry. Thus, in the following, we only discuss the planar features.
KNN is used for each transformed feature point $\mathbf{x}^{L_p} \in \mathbf{F}^{L_p}_{L_{\alpha}}$ to find the $\mathbf{K}$ nearest points $\pi(\mathbf{x}^{L_p})$ in $\mathbf{M}^{L_p}_{L_{o,i}}$.
Then for the planar points, we fit these neighbor points into a plane in $\mathcal{F}_{L_p}$.
The coefficient of a planar point can be solved by the linear equation defined by $\bm{\omega}^T \mathbf{x}' + d = 0, \mathbf{x}' \in \pi(\mathbf{x}^{L_p})$, where $\bm{\omega}$ is the plane normal direction and $d$ is the distance to the origin of $\mathcal{F}_{L_p}$.
We denote $m = \left[\mathbf{x}, \bm{\omega}, d\right] \in \mathbf{m}_{L_{\alpha}}$ for each planar feature point $\mathbf{x} \in \mathbf{F}_{L_{\alpha}}$ as one of the relative lidar measurements.
To be mentioned, in each relative lidar measurement $m \in \mathbf{m}_{L_{\alpha}}$, $\mathbf{x}$ is defined in $\mathcal{F}_{L_\alpha}$, and $\bm{\omega}$ and $d$ are defined in $\mathcal{F}_{L_p}$.

\subsection{Lidar Sweep Matching} \label{sec:lidar_matching}

The relative lidar measurements can provide relative constraints between the $pivot$ lidar pose and the following lidar poses.
Our method optimizes all the poses in the optimization window, including the first pose $\mathbf{T}^{W}_{L_p}$, i.e., $\mathcal{F}_{L_p}$ is not fixed. Thus, each item in the lidar cost function involves the poses of two lidar sweeps, $\mathbf{T}^{W}_{L_p}$ and $\mathbf{T}^{W}_{L_{\alpha}}, \alpha \in \{p+1, \cdots, j\}$. Optimizing the $pivot$ pose will help to minimize the pre-integration error better and ensure the sensor pair align with gravity.
The states we estimate are those of the IMU; thus we need to introduce extrinsic parameters to represent the lidar constraints by IMU states. The relative transformation from the latter lidar poses to the $pivot$ one in the window can be defined as

\begin{equation} \label{eqn:relative_transform}
  \begin{aligned}
    \mathbf{T}^{L_p}_{L_{\alpha}} = {\mathbf{T}^L_{B}\mathbf{T}^W_{B_p}}^{-1} \mathbf{T}^W_{B_{\alpha}}{\mathbf{T}^L_B}^{-1} =
    \begin{bmatrix}
      \mathbf{R}^{L_p}_{L_{\alpha}} & \mathbf{p}^{L_p}_{L_{\alpha}}\\
      \mathbf{0} &          1
    \end{bmatrix}
  \end{aligned}.
\end{equation}

With the previous correspondences, the residual for each relative lidar measurement $m =  \left[\mathbf{x}, \bm{\omega}, d\right] \in \mathbf{m}_{L_{\alpha}}, \alpha \in \{p+1, \cdots, j\}$ can be represented as point-to-plane distance
\begin{equation} \label{eqn:lidar_constraint}
  \begin{aligned}
    \mathbf{r}_{\mathcal{L}}(m, \mathbf{T}^W_{L_p}, \mathbf{T}^W_{L_{\alpha}},  \mathbf{T}^L_B) = \bm{\omega}^T(\mathbf{R}^{L_p}_{L_{\alpha}} x+\mathbf{p}^{L_p}_{L_{\alpha}}) + d.
  \end{aligned}
\end{equation}

\subsection{Optimization}\label{sec:optimization}
To obtain the optimized states, a fixed-lag smoother and marginalization are applied. The fixed-lag smoother keeps $N_s$ IMU states in the sliding window, from $\mathbf{X}^W_{B_p}$ to $\mathbf{X}^W_{B_j}$, as shown in Fig. \ref{fig:local_window}. The sliding window helps to bound the amount of computation. When new measurement constraints come, the smoother will include the new states and marginalize the oldest states in the window. The whole of the states to be estimated in detail is

\begin{equation} \label{eqn:states}
  \begin{aligned}
    \mathbf{X} = \left[\mathbf{X}^W_{B_p}, \cdots, \mathbf{X}^W_{B_j}, \mathbf{T}^L_B \right]
  \end{aligned}.
\end{equation}

Then the following cost funtion with a Mahalanobis norm is minimized to obtain the MAP estimation of the states $\mathbf{X}$,
\begin{equation} \label{eqn:cost_func}
  \begin{aligned}
    \min_{\mathbf{X}} \frac{1}{2} \Bigg\{ &\left\lVert \mathbf{r}_{\mathcal{P}}(\mathbf{X}) \right\rVert^2 + \sum_{\substack{m \in \mathbf{m}_{L_{\alpha}}\\ \alpha \in \{p+1, \cdots, j\}}}{\left\lVert \mathbf{r}_{\mathcal{L}}(m, \mathbf{X})  \right\rVert^2_{\mathbf{C}^m_{L_{\alpha}}}} \\
    &+ \sum_{\substack{\beta \in \{p, \cdots, j-1 \}}}{\left\lVert \mathbf{r}_{\mathcal{B}}(z^{\beta}_{\beta+1}, \mathbf{X})  \right\rVert^2_{\mathbf{C}^{B_{\beta}}_{B_{\beta + 1}}}} \Bigg\}
  \end{aligned},
\end{equation}
where $\mathbf{r}_{\mathcal{P}}(\mathbf{X})$ is the prior items from marginalization. $\mathbf{r}_{\mathcal{L}}(m, \mathbf{X})$ is the residual of the relative lidar constraints and $\mathbf{r}_{\mathcal{B}}(z^{\beta}_{\beta+1}, \mathbf{X})$ is the residual of the IMU constraints. The cost function in the form of a non-linear least square can be solved by the Gauss-Newton algorithm, which takes the form $\mathbf{H}\delta\mathbf{X} = -\mathbf{b}$. We use Ceres Solver \cite{ceres-solver} to solve the problem.

The lidar constraint $\mathbf{r}_{\mathcal{L}}(m, \mathbf{X})$ can be derived from Eq. (\ref{eqn:lidar_constraint}) for each relative lidar measurement. The covariance matrix $\mathbf{C}^m_{L_{\alpha}}$ is determined by the lidar accuracy. Similar to the one in \cite{qin2018vins}, IMU constraint $\mathbf{r}_{\mathcal{B}}(z^{\beta}_{\beta+1}, \mathbf{X})$ can be obtained from the states and IMU pre-integration,  $\left\lVert \mathbf{r}_{\mathcal{P}}(\mathbf{X}) \right\rVert^2= \mathbf{b}^T_{\mathcal{P}} \mathbf{H}^+_{\mathcal{P}} \mathbf{b}_{\mathcal{P}}$ can be obtained by Schur complement (detailed in \cite{ye2018supplementary}).

\section{Refinement with Rotational Constraints} \label{sec:mapping}

\begin{figure}[tb]
\centering
\includegraphics[width=0.4\textwidth]{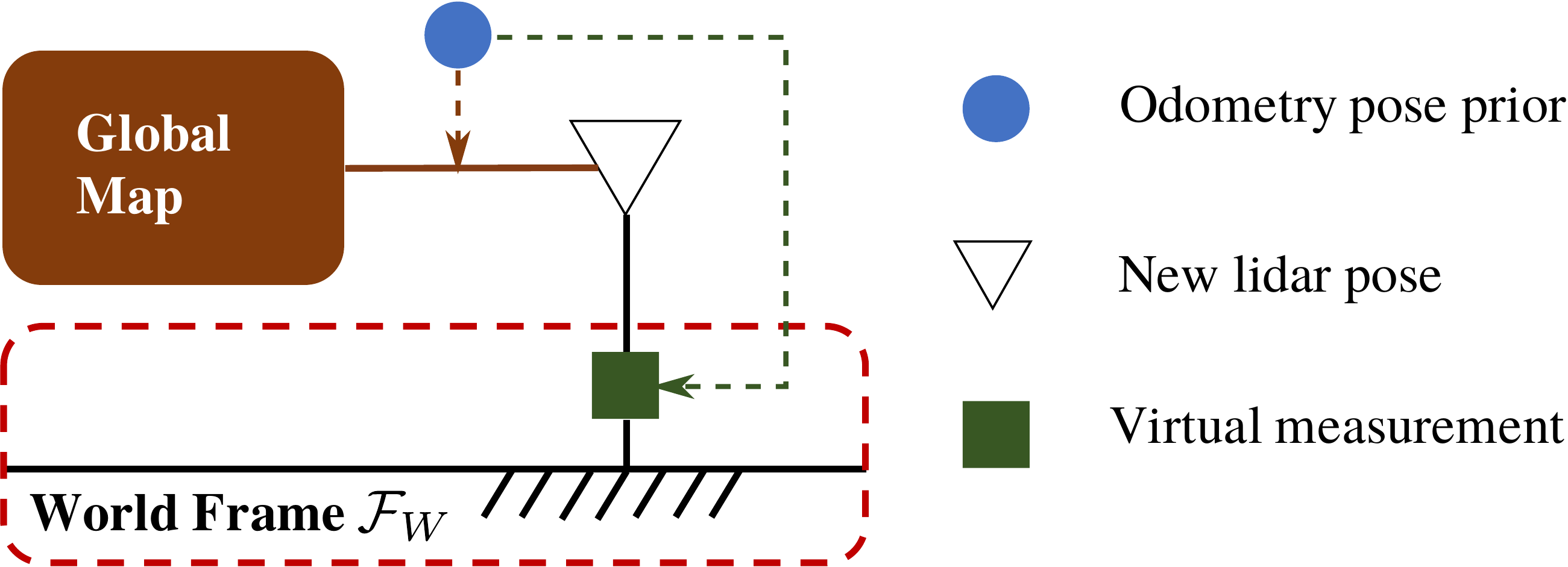}
\caption{Rotationally constrained mapping. The odometry pose first serves as a prior for the global point-cloud registration. Then the rotational component from the odometry is applied as a virtual measurement to constrain the optimization.}
\label{constrained_mapping}
\end{figure}

Registering the feature points to a global map, instead of local maps, can constrain the lidar poses to a consistent world frame $\mathcal{F}_W$.
Our refinement method uses the relative lidar measurements $\mathbf{m}_L$ as the ones in Sec. \ref{sec:odom}. Since the global map is a by-product of the refinement, we also refer to it as a mapping method.
A cost function to align the latest lidar feature points with the global map can be formed as
\begin{equation} \label{eqn:map_constraint}
  \begin{aligned}
    &\mathbf{C}_{\mathcal{M}} = \sum_{m \in \mathbf{m}_L} \left\lVert \mathbf{r}_{\mathcal{M}}(m, \mathbf{T}^W_{L}) \right\rVert^2\\
    &\mathbf{r}_{\mathcal{M}}(m, \mathbf{T}) = \bm{\omega}^T({\mathbf{R}}\mathbf{x} + \mathbf{p}) + d
  \end{aligned},
\end{equation}
where $\mathbf{T} = \mathbf{T}^W_L$ is the latest to-be-estimated lidar pose, and $m$ is the relative lidar measurement with the feature point $\mathbf{x}$ in $\mathcal{F}_L$ and the coefficients $\bm{\omega}, d$ defined in $\mathcal{F}_W$.
Then we can use a similar Gauss-Newton method to minimize $\mathbf{C}_{\mathcal{M}}$. The optimization is carried out by the residual $\mathbf{C}_{\mathcal{M}}$ and Jacobians $\mathbf{J}^{\mathbf{C}}_{\mathbf{p}}$ and $\mathbf{J}^{\mathbf{C}}_{\bm{\theta}}$, where $\bm{\theta}$ is the error state of the corresponding quaternion $\mathbf{q}$.
However, with the accumulated rotation error and after long-term operation, the merged global map cannot align with gravity accurately.
This can lead further mapping to wrongly align with a tilted map.
Inspired by \cite{zheng2018odometry} which optimizes SE(3) with SE(2)-constraints, we propose a constrained mapping strategy.
This strategy utilizes the rotational constraints from the lidar-IMU odometry, which ensures the final map always aligns with gravity. Fig. \ref{constrained_mapping} illustrates the structure of the rotationally constrained mapping.

Given the property that the orientation along the $z$-axis has higher uncertainty, and that the other two DoF of the orientation are much more close to the true value, we can constrain the cost function by modifying the Jacobian of the orientation as (detailed derivations in \cite{ye2018supplementary}),
\begin{equation} \label{eqn:modified_ori}
  \begin{aligned}
    &\mathbf{J}^{\mathbf{C}}_{\bm{\theta}_z} = \mathbf{J}^{\mathbf{C}}_{\bm{\theta}} \cdot (\breve{\mathbf{R}})^T \cdot \breve{\mathbf{\Omega}}_z \\
    &\breve{\mathbf{\Omega}}_z = \begin{bmatrix}
      \epsilon_x & 0 & 0 \\
      0 & \epsilon_y & 0 \\
      0 & 0 & 1
    \end{bmatrix}
  \end{aligned},
\end{equation}
where $\breve{(\cdot)}$ denotes the estimation of the state in the last iteration, and $\breve{\mathbf{\Omega}}_z$ is an approximiation of the information matrix of the orientation w.r.t $\mathcal{F}_W$, and $\epsilon_x$ and $\epsilon_y$ can be obtained by the information ratio of the $x$- and $y$-axes orientation to the $z$-axis orientation in $\mathcal{F}_W$.

After that, we use $\mathbf{J}^{\mathbf{C}}_{\mathbf{p}}$ and $\mathbf{J}^{\mathbf{C}}_{\bm{\theta}_z}$ as the Jacobians instead, and these are needed for the optimization step. The incremental lidar poses can be obtained as $\delta \bm{\theta}_z$ and $\delta \mathbf{p}$, which lead to the updated lidar states $\tilde{\mathbf{p}}$ and $\tilde{\mathbf{q}}$
\begin{equation} \label{eqn:modified_update}
  \begin{aligned}
    &\tilde{\mathbf{p}} = \breve{\mathbf{p}} + \delta \mathbf{p} \\
    &\tilde{\mathbf{q}} = \begin{bmatrix} \frac{1}{2}\delta \bm{\theta}_z \\ 1 \end{bmatrix} \otimes \breve{\mathbf{q}}
  \end{aligned}.
\end{equation}

\section{Implementation} \label{sec:Implemetation}
 Different sensor configurations, system initialization and different parameters for indoor and outdoor tests are introduced in this section.
\subsection{Different Sensor Configurations}
Sensor pairs can be configured differently.
For a hand-held sensor pair, such as the one in Fig. \ref{fig:indoor_sensors}, the lidar and IMU are close to each other.
Thus the pipeline remains the same as what was introduced before. But for sensor pairs mounted on cars, the two sensors are usually farther away from each other.
For example, Fig. \ref{fig:golf_car} shows an IMU mounted above the car's base link, while the lidar is mounted at the front of the car.
Instead of auto-calibrating all translation parameters, a prior item for the extrinsic translational parameters is added to Eq. (\ref{eqn:cost_func}) for the tests on the cars.

\subsection{Initialization}

At first, there are no estimated poses for the lidar feature points.
Therefore, roughly accurate matching algorithms are needed.
We adopt \cite{zhang2014loam}'s lidar odometry in the initialization step.
With the provided lidar poses, sufficient motions of the sensor pair are required to make the IMU states observable \cite{jones2011visual}.
Then the poses of the lidar and the IMU measurements are used to initialize the IMU states, which can be solved by the methods introduced in \cite{qin2018vins} and \cite{mur2017visual}. For the tests in Sec. \ref{sec:results}, we follow the initialization method in \cite{qin2018vins}, which also linearly initializes the extrinsic parameters. Next, with the initial states and the newly coming measurements, the non-linear optimization within local windows will be carried out iteratively to estimate the states.

%% file: sections/5_results.tex
\section{Tests and Analyses} \label{sec:results}

Indoor and outdoor tests are conducted to evaluate our method. The quantitative and qualitative results are provided in the following sections.

\subsection{Quantitative Analysis}

\begin{figure}
  \centering

  \begin{subfigure}[b]{0.23\textwidth}
    \centering
    \includegraphics[width=\textwidth]{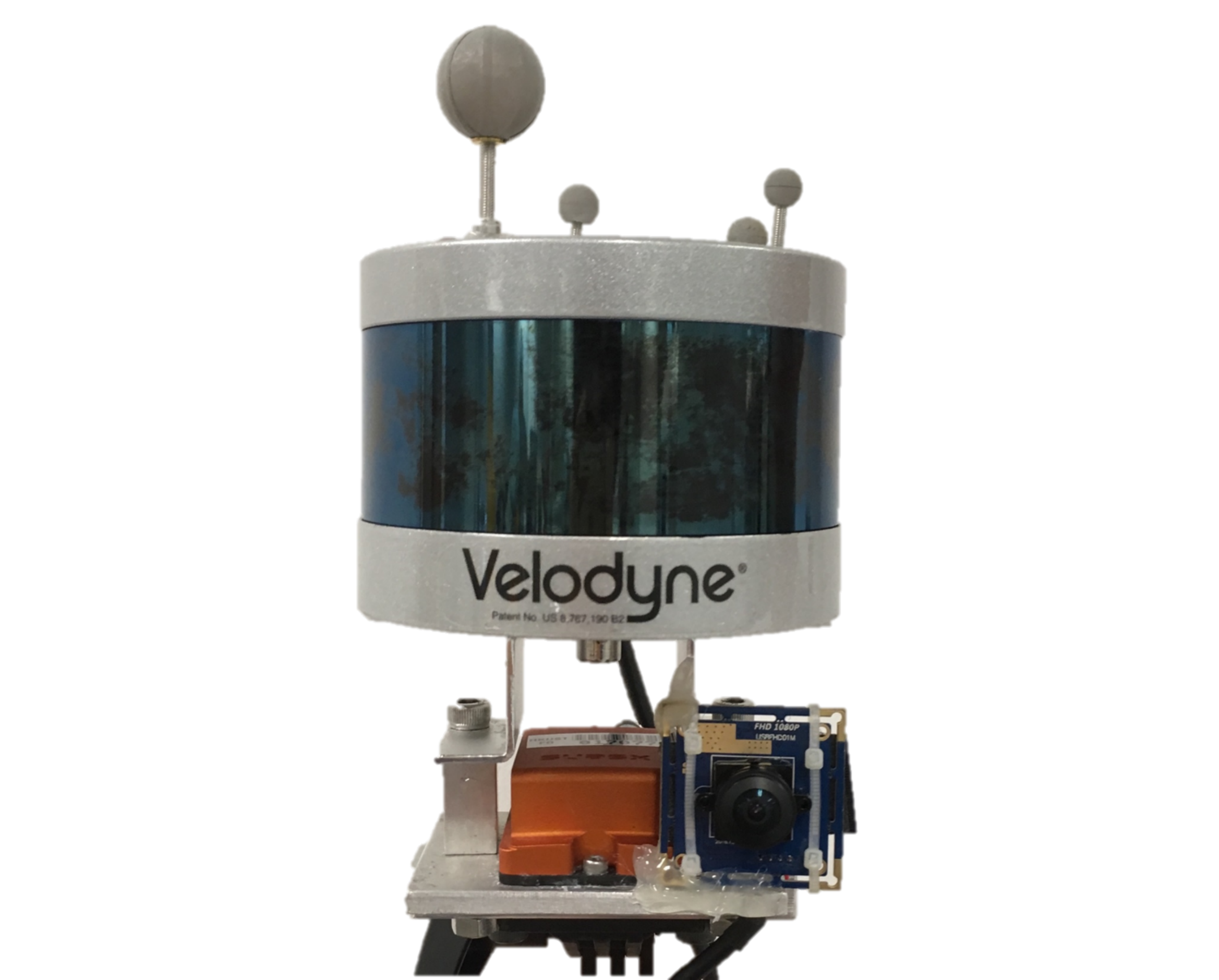}
    \caption{}
    \label{fig:indoor_sensors}
  \end{subfigure}
  \begin{subfigure}[b]{0.23\textwidth}
    \centering
    \includegraphics[width=\textwidth]{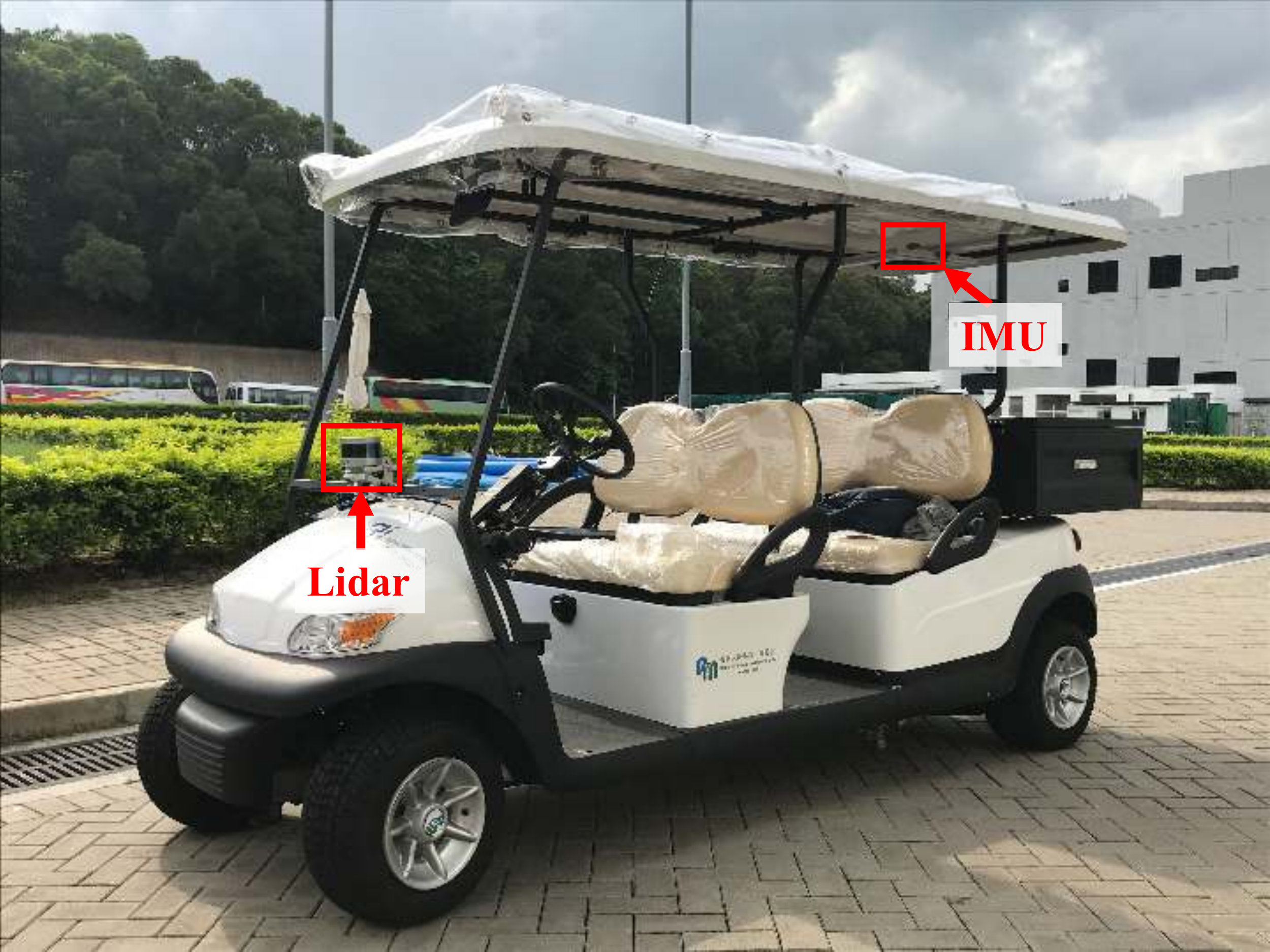}
    \caption{}
    \label{fig:golf_car}
  \end{subfigure}

  \caption{(a) The sensor configuration for quantitative analysis and indoor tests.
    Note that the attached camera is only used to record the test scenes.
    (b) Outdoor golf car configuration.
  }
  \label{fig:sensors}
  \vspace{-0.5em}
\end{figure}

To quantitatively analyze our method, the sensor pair shown in Fig. \ref{fig:indoor_sensors} is used. A Velodyne VLP-16 lidar with 16 lines is mounted above an Xsens MTi-100 IMU. The reflective markers can provide the ground-truth poses using the motion capture system. The lidar is configured to have a 10Hz update rate, and IMU updates at 400Hz. The estimated trajectories from different methods are aligned with the ground-truth using \cite{umeyama1991least}.

\subsubsection{Tests under Different Motion Conditions}

\begin{table}[]
  \centering
  \caption{Translational and rotational errors w.r.t. ground-truth.
    The motions in the 6 sequences vary from fast to slow.
  }
  \label{tab:vicon_results}

  \begin{tabular}{@{}l|l|l|l|l|l|l@{}}
    \toprule
    Errors & Sequence & LOAM & \begin{tabular}[c]{@{}l@{}}LIO-\\ raw\end{tabular} & \begin{tabular}[c]{@{}l@{}}LIO-\\ no-ex\end{tabular} & LIO & \begin{tabular}[c]{@{}l@{}}LIO-\\ mapping\end{tabular} \\ \midrule
    \multirow{6}{*}{\begin{tabular}[c]{@{}l@{}}Trans-\\ lation\\ RMSE \\ (m)\end{tabular}}
           & fast 1 & 0.4469 & 0.2464 & 0.0957 & \textbf{0.0949} & \textbf{0.0529} \\
           & fast 2 & 0.2023 & 0.4346 & 0.1210 & \textbf{0.0755} & \textbf{0.0663} \\
           & med 1 & 0.1740 & 0.1413 & 0.1677 & \textbf{0.1002} & \textbf{0.0576} \\
           & med 2 & \textbf{0.1010} & 0.2460 & 0.3032 & 0.1308 & \textbf{0.0874} \\
           & slow 1 & \textbf{0.0606} & 0.1014 & 0.0838 & 0.0725 & \textbf{0.0318} \\
           & slow 2 & \textbf{0.0666} & 0.1016 & 0.0868 & 0.1024 & \textbf{0.0435} \\ \midrule
    \multirow{6}{*}{\begin{tabular}[c]{@{}l@{}}Rotation\\ RMSE \\ (rad)\end{tabular}}
           & fast 1 & 0.1104 & 0.1123 & 0.0547 & \textbf{0.0545} & \textbf{0.0537} \\
           & fast 2 & 0.0763 & 0.1063 & 0.0784 & \textbf{0.0581} & \textbf{0.0574} \\
           & med 1 & 0.0724 & 0.0620 & 0.0596 & \textbf{0.0570} & \textbf{0.0523} \\
           & med 2 & 0.0617 & 0.0886 & 0.0900 & \textbf{0.0557} & \textbf{0.0567} \\
           & slow 1 & \textbf{0.0558} & 0.0672 & 0.0572 & 0.0581 & \textbf{0.0496} \\
           & slow 2 & 0.0614 & 0.0548 & 0.0551 & \textbf{0.0533} & \textbf{0.0530} \\ \bottomrule
  \end{tabular}

\end{table}

Table \ref{tab:vicon_results} shows the root mean square error (RMSE) results under different motion speeds and different methods, where LOAM \cite{zhang2014loam} is regarded as the baseline.
LIO is our local window optimized odometry method.
LIO-raw and LIO-no-ex are the same as LIO expect that the motion compensation or the online extrinsic parameter estimation is cut off, respectively.
LIO-mapping is from the results of the mapping with rotaional constraints.
The two best results are shown in bold.

From the results, we see that LIO-mapping can always provide accurate estimation of translational (position) and rotational (orientation) states in all cases. LIO has better performance when motion is faster, which produces more IMU excitation. But it suffers from drift if the motion is slow, since the local map is relatively sparse at this time. The table also shows that with motion compensation and online extrinsic parameter estimation, LIO can provide better performance, especially when motions are rapid.

\subsubsection{Tests of Drift over Time}

\begin{figure}[!ht]
  \centering
  \includegraphics[width=0.35\textwidth]{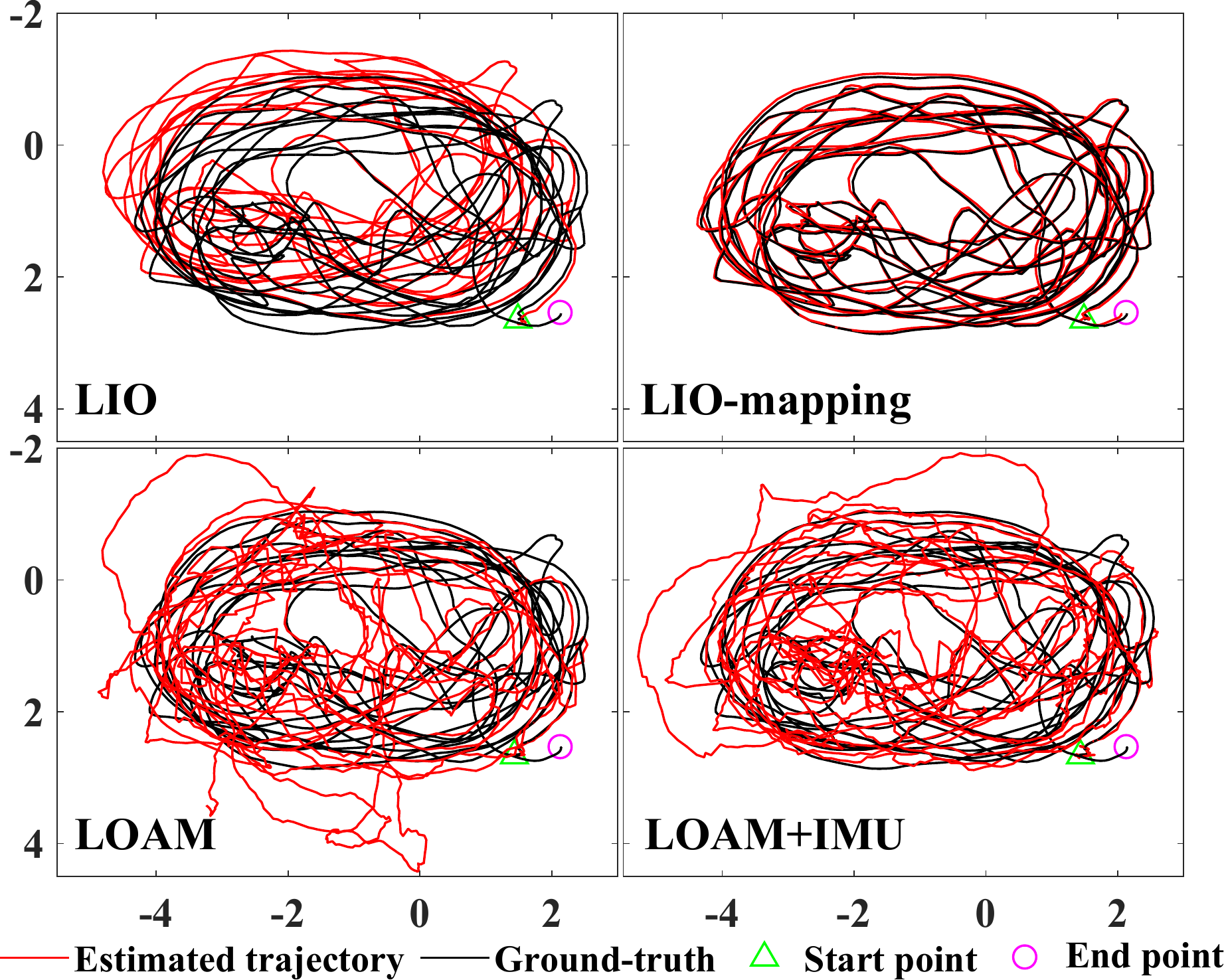}
  \caption{Trajectories from different methods. LIO can provide relatively accurate poses. Due to the small local window it uses, it has drift when run long-term. LIO-mapping can eliminate the drift with the help of a consistent map. Neither LOAM nor LOAM+IMU can work when the motion becomes rapid (in the latter half of the test).}
  \label{trajectory}
  \vspace{-1.5em}
\end{figure}

\begin{figure}[!ht]
  \centering

  \begin{subfigure}[b]{0.4\textwidth}
     \begin{minipage}[b]{0.05\linewidth}
       \caption{
       }
    \end{minipage}
    \begin{minipage}[b]{0.9\linewidth}
      \includegraphics[width=\textwidth]{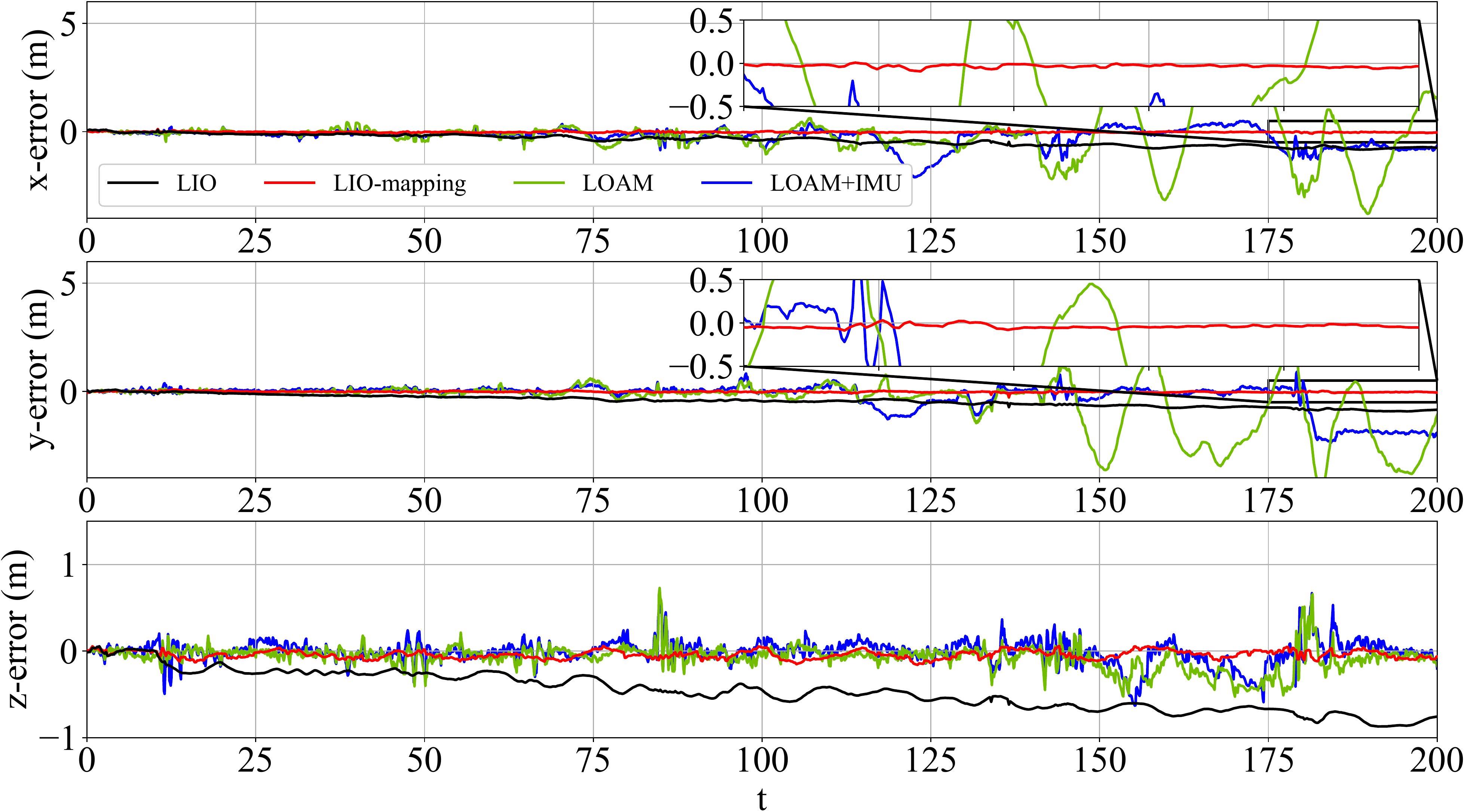}
    \end{minipage}
    \label{fig:xyz_errors}
  \end{subfigure}
  \begin{subfigure}[b]{0.4\textwidth}
    \begin{minipage}[b]{0.05\linewidth}
      \caption{
      }
    \end{minipage}
    \label{fig:ypr_errors}
    \begin{minipage}[b]{0.9\linewidth}
      \includegraphics[width=\textwidth]{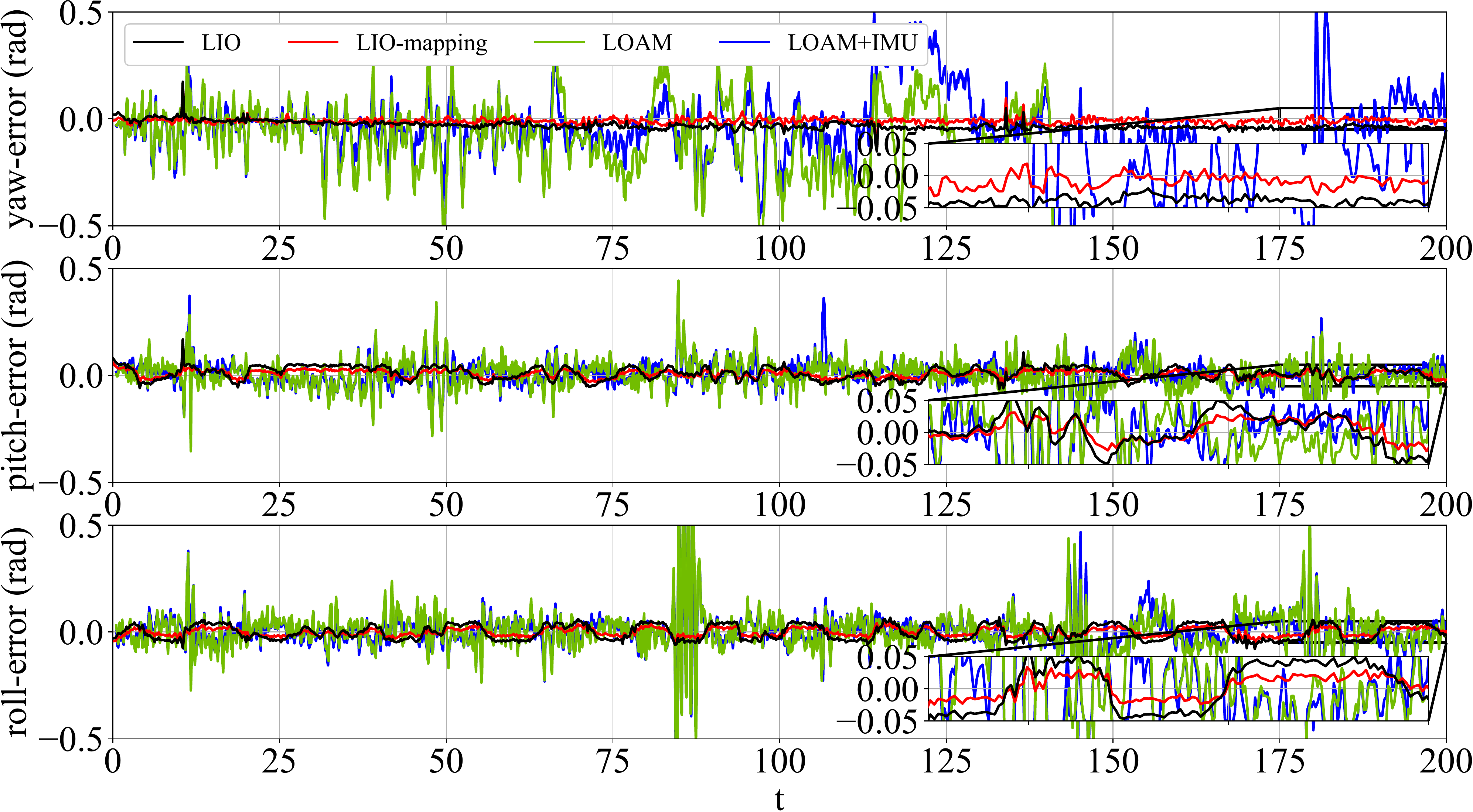}
    \end{minipage}
  \end{subfigure}

  \caption{%
    (a) Translation errors. (b) Rotational errors. In general, our methods (both LIO and LIO-mapping) can provide much smoother results than their counterparts.
  }
  \label{fig:xyz_ypr_errors}
  \vspace{-0.8em}
\end{figure}

To evaluate how the error changes with time, we test the algorithms in a longer test. The first 50 estimated poses are aligned with the ground-truth.
The final trajectories from the different methods are shown in Fig. \ref{trajectory}, and the translational and rotational errors are shown in Fig. \ref{fig:xyz_ypr_errors}. The results show that LIO can provide relatively accurate poses and constrain $roll$ and $pitch$ close to the ground-truth, but it suffers from drift.
Neither the method without IMU fusion (LOAM) nor the one with loosely coupled fusion (LOAM+IMU) can provide robust estimation when the motion becomes rapid (in the latter half of the test).
LIO-mapping benefits from rotational constraints provided by LIO, and further registers the current sweep to the global map. Thus, it results in less drift of the trajectory and greater consistency of the state estimation.

\subsection{Qualitative Results}

Several tests with different sensor configurations and environments are carried out in order to show the improvements in challenging scenarios, including indoor hand-held and outdoor campus golf cart tests, with the configurations as shown in Fig. \ref{fig:sensors}, and tests on the KAIST Urban dataset \cite{jeong2018complex}. Due to the limited space, these pose estimation and mapping results are shown in \href{https://sites.google.com/view/lio-mapping}{the supplementary video}.

\subsection{Running Time Analysis}

We run this test with an Intel i7-7700K CPU at 4.20GHz, 16GB RAM.
The lidar intervals
vary indoors (0.2s) and outdoors (0.3s), depending on the number of feature points in a sweep (typically more feature points outdoors, around 3000, than indoors, 1000).
These intervals help to build larger maps and skip some of the lidar sweeps to fulfill real-time computation.
The mean running time of our method can be found in Table \ref{tab:running_time} using the data from a 16-line 3D lidar.
The time stands for the processing time of each of the new inputs for a module, i.e., raw IMU measurements, lidar measurements and odometry outputs.
Note that the odometry and mapping are in different threads.
The mapping thread processes the outputs from the odometry thread.
The prediction of the IMU is operated based on the optimized states, alongside the optimization.
Thus, it can run as fast as the output rate of the IMU.

\begin{table}[]

\centering
\caption{Mean running time analyses for 16-line 3D lidar.
}
  \label{tab:running_time}

  \begin{tabular}{@{}l|lllll@{}}
    \toprule
    \multirow{2}{*}{Scenarios} & \multicolumn{3}{c}{Time ($ms$)}                                                                                                                \\ \cmidrule(l){2-4}
                               & \multicolumn{1}{l|}{Prediction} & \multicolumn{1}{l|}{Odometry} & Mapping \\ \midrule
    Indoor                     & \multicolumn{1}{l|}{0.0127}      & \multicolumn{1}{l|}{128.7}       & 108.3   \\ \midrule
    Outdoor                    & \multicolumn{1}{l|}{0.0102}      & \multicolumn{1}{l|}{213.5}       & 167.6    \\ \bottomrule
  \end{tabular}
  \vspace{-0.3em}
\end{table}